# The Effects of using created Synthetic images in Computer Vision training

John W. Belanger Smutny
Software Engineer
Virginia Tech
Arlington, VA, USA
smutnyj@vt.edu

Creed Jones, PhD
Computer Engineering
Virginia Tech
Blacksburg, VA, USA
crjones4@vt.edu

*Abstract*— **This paper investigates how rendering engines, like Unreal Engine 4 (UE), can be used to create synthetic images to supplement datasets for deep computer vision (CV) models in image abundant and image limited use cases. Using rendered synthetic images from UE can provide developers and businesses with a method of accessing nearly unlimited, reproducible, agile, and cheap training sets for their customers and applications without the threat of poisoned images from the internet or the cost of collecting them. The validity of these generated images are examined by testing the change in model test accuracy in two different sized CV models across two binary classification cases (Cat vs Dog and Weld Defect Detection). In addition, this paper provides an implementation of how to measure the quality of synthetic images by using pre-trained CV models as auditors.**

**Results imply that for large (VGG16) and small (MobileNetV3-small) parameter deep CV models, adding >60% additional synthetic images to a real image dataset during model training can narrow the test-training accuracy gap to ~1-2% without a conclusive effect on test accuracy compared to using real world images alone. Likewise, adding <10% additional real training images to synthetic only training sets decreased the classification error rate in half, then decreasing further when adding more real training images. For these cases tested, using synthetic images from rendering engines allow researchers to only use 10% of their real images during training, compared to the traditional 50-70%. This research serves as an example of how to create synthetic images, guidelines on how to use the images, potential restrictions and possible performance improvements for data-scarce projects.** (*Abstract*)

*Keywords— Convolutional Neural Networks, synthetic image generation, image rendering, dataset creation.*

## I. INTRODUCTION

All machine learning (ML) models require large amounts of training data to learn from. Gathering a large enough real dataset with enough image variation can be difficult depending on the application. Throughout a CV application's life cycle, training, testing and retraining the model requires large amounts of varied images. Images taken from multiple contexts, camera angles and lighting contribute to expanding a class's feature space. If a dataset is not publicly available or is for a niche application, gathering data can require significant time (manually collecting each image), financial expenditure (licensing), risk (dataset poisoning), and/or legal constraints (copyright of images from the internet). Even already gathered images could have access limits, usage constraints, or limited terms of use. One alternative is to generate photo-realistic images using free graphics rendering software already used in TV, film, video games, and various business applications. Generating synthetic data in visualization tools such as Epic Game's Unreal Engine (UE) can provide teams with a near infinite supply of images that replicate objects of interest in a controlled and repeatable environment. These virtual environments can be changed to follow new customer requirements without the need to re-gather massive datasets. Teams can even model situations that haven't been seen before, thus increasing the model's flexibility and reducing the need for model re-training.

This research defines a 'synthetic' image as any 2D image created from a rendering software in a digital environment. Computer graphics over the years have improved in its ability to simulate the real world. All the objects in said renderings are digitally created via a mesh & textures (similar to CAD models). One common method of creating this type of image is to use tools such as the Epic Game's Unreal Engine, Microsoft's Unity Engine, or another rendering software. 'Synthetic' images do not include scanned copies of photographs or images created by Generative-Adversarial Networks (GANs). Images taken by traditional cameras are digitized approximations of the real world based on the camera's resolution. These images are real objects, synthetic images are simulations of objects. Generative images are created from learned statistical approximations that the GAN was trained on. Unless the developer controls the model or compiler seed, these statistical approximations are not commonly reproducible. Also, synthetic images are different from image augmentation (flips, sheer, blur) since they are rendered using simulation software rather than modifying an already existing image. You can augment a synthetic image, but you cannot make a traditionally gathered image synthetic. This research focuses on images that designers can intentionally and repeatably create on their own for their own specific application.

## II. RELATED RESEARCH

Starting as early as 2016 several researchers, such as Dr. Thornström of Linkoping University and Dr. Gastelum of Sandia National Labs, have explored how rendered images can improve CV model training. They proposed mixing real images with their created synthetic images to improve classifier accuracy. Dr Thornström used simulations to place objects in unique situations, while Dr. Gastelum tried to mimic their use case's industrial environment. Both outlined how synthetic images could improve classifier accuracy, but Dr. Gastelum's work was limited by the lack of simulation variation [12][4]. Other additional researchers explored the uses of synthetic images in CV training [3]. An academic survey created by Dr. Man further summarized how these rendered images could be used for individual projects and definitions for the field [7]. Meanwhile, Dr. Qiu's *UnrealCV* python module improved accessibility for users to create their own rendered images for ML training [10][2]. This paper methodologies draw inspiration from Dr. Thornström's procedure and Dr. Qiu's coding framework. Specifically exploring the benefit of synthetic images, how they can replace a dataset's portion of real images and expands upon their work by providing a method to measure how close rendered images are to their real-world counterparts using pre-trained classifiers.

## III. METHODOLOGY

To understand how mixed training sets affect a CV model's performance, this research repeatedly trains CV models with different mixtures of real and synthetic images (see Fig. 3.1). One image type is held constant while the other increases (starting at zero). Specifically, training CV models on a training set with 1) a constant amount of real images & variable synthetic images and 2) variable amounts of real images & constant amounts of synthetic images. By steadily increasing the number of one image type, the resulting model accuracy and loss metrics directly show how the training set mixture affects the resulting model. Keep in mind, in all outlined experiments, synthetic images are only used during model training. All model validation and testing sets contain only real images.

| id | CV Architecture | # of Real training images | # of Synthetic training images |
|---|---|---|---|
| 1 | VGG16 | constant | variable |
| 2 | VGG16 | variable | constant |
| 3 | MobileNetV3-Small | constant | variable |
| 4 | MobileNetV3-Small | variable | constant |

Fig 3.1. Outline of training mixes tested.

### A. Datasets used

Two binary classification tasks are explored in this paper; 1) is the image of a 'cat or dogs' and 2) (product quality) identifying if a metal weld is defective or not. Each classification task uses a unique set of real and synthetic images. The Cat_Dog real dataset comes from the ImageNet classification standard set (a balanced set of 25,000 images) [17], while the synthetic dataset is a balanced set of 47864 rendered images from publicly available cat & dog models. The Welding real image dataset contains 2221 images from several public repositories [14][15][16][18][19]. Only 38.2% of the initially found images are used due to image duplication, improper framing, and irrelevance to metal welding. For simplicity, only 2000 real images are used in this research. Four groups make up the 6528 synthetic images: *No change*, *porosity defect*, *crack defect*, and *incomplete weld defect*. Both real and synthetic welding datasets are imbalanced at a rate of 7 Defects to 3 Good welds.

| Use Case | Range of Real Training images used | Range of Synthetic Training images used | Number of Real Validation images used | Number of Real Test images used |
|---|---|---|---|---|
| Cat vs Dog | [0, 6000] | [0, 6000] | 2000 | 2000 |
| Weld Defect Detection | [0, 1200] | [0, 1200] | 400 | 400 |

Fig. 3.2. Dataset split breakdown for training, validation, and test sets during model training. Only the training set contains synthetic images. Validation and Test sets only use real images.

An additional synthetic class (*minor imperfections*) was created to consider minor surface variations in a weld while still being non-defective (*Good*). This class, and its created images, were removed from the synthetic training set because during testing the class did not appear to improve model accuracy or generalization. It is believed that the *minor imperfection* class blurred the feature spaces of the *Defect* and *Good* classes, reducing test accuracy.

### B. Creating Synthetic Images

All synthetic images are collected from executing a python script in a UE Editor instance. During execution, the python script runs on the UE Editor's thread via callback, sending commands to the Editor to render each image. Every 30 frames a sequence of commands is sent to change the lighting, background, and camera angles (resulting in 80 unique ~1980x1240 screenshots a minute). This rate is chosen based on used hardware and time required each callback to execute all necessary UE commands prior to prompting the next callback.

The Cat_Dog synthetic dataset is created from 13 Cat and 10 Dog publicly available low-polygon (~10,000) .fbx model files [1]. The screenshots capture 100 different camera angles, 10 backgrounds, and 8 lighting settings. For the welding case, the 8160 synthetic weld images are collected from 6 color variations (aluminum, steel, copper, gold, etc), 68 unique locations along a random weld seam (provided by a purchased UE material [20]. At each point, 4 screenshots are taken at random camera angles and point light offsets of each of the five seam modifications described in the previous section. Examples are included on the next page (Fig 3.3 & Fig 3.4).

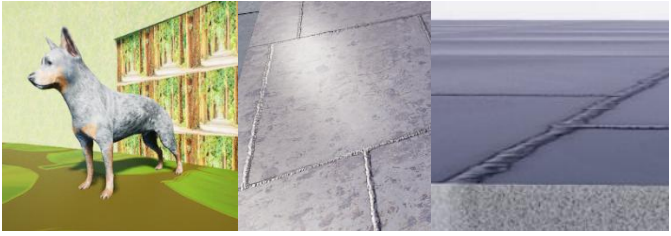

Fig. 3.3. Examples of images used in the (left) Cat_Dog dataset and (middle & right) Weld Defect datasets. In the Right image, that the 'weld' not 3-dimensional. Any perceived height is an illusion during rendering.

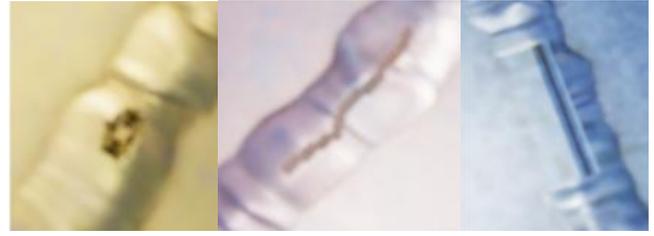

Fig. 3.4. Examples of (left) porosity, (middle) crack, and (right) incomplete weld defects created from Normal, Metallic, Roughness, and Ambient Occlusion maps purchased from the UE Store [20].

The three weld defects (*porosity*, *crack*, & *incomplete weld*) are simulated by modifying UE Texture Maps that combine to form the defect captured in each screenshot. Each defect is applied to a ~34 pixels wide weld seam with different dimensions at each of the 68 locations. Each defect's dimensions are randomly chosen from a range of set values. *Porosity* defects are formed by outlining an ellipsis then editing the Normal and Ambient Occlusion (AO) maps. *Cracks* are formed from offsetting a line segment. *Incomplete welds* consist of removing the Normal & AO map from parts of the weld seam and inserting a black box representing the seam in between the metal plates. As stated previously, the *minor imperfections* class was removed but was originally created to reflect small changes to the weld but not enough to be a Defect. For this project, {*no change*} are classified as Good welds (positives - 1), while {*porosity*, *crack*, & *incomplete welds*} are classified as Defects (negatives – 0).

### C. Models Implemented and Training

VGG16 and MobileNetV3-small are the two models examined in this research, one large and one small. The VGG16 model is a model with many parameters that are best run where there are minor limitations on power consumption and computer memory, such as hosting on an on-site or cloud server. The model has a common convolutional neural network architecture involving 16 sets of *Convolution-Pooling-BatchNorm* layers [11].

MobileNetV3-small is a variation of the MobileNetV3 model that has the fewest number of trainable parameters. The model is specifically designed for running on mobile devices & on-chip hardware, prioritizing lower power usage and low memory footprint for a reduced latency from model input to output. MobileNetV3's architecture utilizes Squeeze-Excite layers that advance upon residual neural networks introduced by Dr Kaiming He's ResNet model [5]. After each convolutional layer, MobileNetV3 uses Squeeze-Excite residual network blocks to pool (squeeze the output) the Conv2D layer's output, passes them through fully connected (FC) layers (excite), before scaling the output for the next Conv2D layer. These FC layers apply weights to each of the previous Conv2D layer filters, thus prioritizing which filters contain the most important features [6]. Fig. 3.5 outlines the differences between the two models.

|  | **VGG16** | **MobileNetV3** |
|---|---|---|
| General Use Case | Minor resource constraints | Constraints on Memory, Power & Latency |
| General Architecture | Conv-AvgPool-BatchNorm | Squeeze & Excite |
| Total Parameters in the published model for 1000 class classification | 138,357,544 (100%) | 2,554,968 (100%) |
| Total Parameters in the transfer learning models used in this research for binary classification | 21,203,521 (15.3%) | 1,365,617 (53.4%) |
| File size of weights (.h5) | 80.9 MB | 5.4 MB |

Fig. 3.5. Summary of the two models used in this report. The models used are significantly smaller than the models in the original paper [10][6].

To prevent overfitting, each model is trained with initial learning (five epochs), learning rate decay (0.95 decay per epoch), early stopping (0.001 tolerance to Val_Acc or Val_ROC), and 2D image augmentations such as flips, shears, zooms, rotation, color contrast, and width+height shifts. Since every training image of every epoch is slightly different, error gradient updates are also different. As a reminder, in all cases, only the training set is augmented. The validation and testing sets are composed of only real images with no augmentations.

### D. Judging Quality of Synthetic Images

Theoretically, if generated synthetic images contain the exact features of real images, then the images would be a perfect simulation of the real world. Such a perfect replication of the world's feature space could remove any need for image collection, but also would prompt a philosophical debate on whether human society is in fact a simulation. This research acknowledges that current tools do not reach that scale, but fear not, this paper's results imply that a perfect reproduction is not necessary to improve CV classification performance.

Pre-trained multi-classification VGG16 and MobileNetV3-small models with ImageNet 1000-class weights are used to judge if real & synthetic images of cats and dogs contain cats or dogs. A successful prediction is if any of the relevant ImageNet cat/dog classes are in the model's top-3 confidence scores. The confidence scores of the correct classifications are used to create a confidence histogram characterizing the real or synthetic data. If both histograms have similar distributions, then the pre-trained model auditor implies there is minimal feature space difference between real and synthetic images.

## IV. EXPERIMENTAL RESULTS

The methods and techniques outlined in the previous section are combined to understand the dynamic between mixed training sets and ways to define the real-synthetic domain gap. It is logical that training on data like your test data yields the best results, but how much of that real data is necessary to train on to benefit your CV model? Does a large fraction of the training set need to be real images or can there only be a couple of samples? This section goes over results that show that only small amounts of real images are needed to significantly improve classifier test accuracy and that only a small amount of rendered synthetic images can enhance classifier performance.

### A. Training models with real & synthetic images

CV model test accuracy is recorded when trained with different compositions of training data. Fig. 4.1 & Fig. 4.2 report the test accuracy gains when additional data is added to the training set. For the use cases tested, Fig. 4.1 outlines how many real images per 100 synthetic images are needed in the training set on to reach maximum test accuracy. Fig. 4.2 outlines the opposite, presenting how many synthetic images per 100 real training images are needed to reach peak test accuracy. Both figures provide examples of the cost-benefit analysis of how many additional images are needed to reach a potential peak in model performance.

The results for measuring model test accuracy with different training set mixes show that synthetic images can replace some real images in a training set. Fig. 4.2 shows that for most of the experiment's cases, fewer additional synthetic training images are needed to reach maximum accuracy and only produce minor benefits for large CV models (compared to when adding real images which provide significant benefit - Fig 4.1). This follows the obvious intuition that real images have similar features to the real world compared to synthetic samples. However, the key is that synthetic images can help reach peak test accuracy with fewer real samples. Fig 4.1 and the weld case in Fig 4.2 show that it's possible to reach the approximate peak test accuracy of exclusively real image trained models with 13-40% fewer real samples when supplemented with synthetic data. Of which, the smaller MobileNetV3-small model saw the greatest reduction in real training data needed to reach similar performance; needing approximately 40% less real images (60 real images per every 100 synthetic images) for a 1.8% & 2.8% drop in test accuracy for their respective cases.

Expanding on the cost-benefit of mixing training sets, general trends emerge when mixing various amounts of each data type. When training CV models with a variable number of real images and constant synthetic images, resulting data graphs show a 1) significant increase in test accuracy and 2) decrease in test-training accuracy gap (implying improved generalization). Both observations hold for both models and both use cases tested. As seen in Fig. 4.3 on the next page, most of the test accuracy improvement occurs with a minimal number of added real images (<300 for these cases) then approaches convergence. Specific to the Weld Detection case, the model's end test accuracy surpassed the training accuracy as the number of real training images increased to the maximum tested. Peak test accuracy occurs when 60-80 real images per 100 synthetic images are added to the training set. Meanwhile, the difference between test accuracy and training accuracy peaks at different points depending on the model and use case (30-80 per 100). Best case, max test accuracy and max test-training accuracy gap occurs at the same ratio to signify the ideal training set balance of real & synthetic images.

When training both CV models with a consistent number of real images and variable number of synthetic images, results show 1) no consistent change in test accuracy and 2) significant decreases in the training-test accuracy gap (implying improved generalization). As seen in Fig. 4.4 on the next page, the VGG16 model's test accuracy reacted inconsistently to the increasing number of synthetic images in the training set, while the MobileNet model showed a slight decrease in test accuracy or no change. Similarly, to the previous 'constant Synthetic' experiment, both models for both use cases exhibited their peak test-training accuracy gap around 60-80 added synthetic images per 100 real training images. This implies that the synthetic images improved the model's generalization and reduced overfitting. However, with real images, max test accuracy was more likely to occur with less added synthetic training images.

| Model | Test Accuracy (synthetic only) | Non-zero # of real training images needed per 100 synthetic images at max Test Accuracy | Max Test Accuracy (mixed) |
|---|---|---|---|
| Cat_Dog | | | |
| VGG16 | 96.4% | 67 | 96.5% |
| MobileNet | 64.1% | 83 | 89.1% |
| Weld | | | |
| VGG16 | 61.4% | 60 | 85.0% |
| MobileNet | 40.2% | 60 | 81.4% |

Fig. 4.1. Effect of real images being introduced into model training sets. Both VGG16 use cases and the Cat_Dog case for MobileNetV3-small display peak performance at the same proportion of real & synthetic training images.

| Model | Test Accuracy (real only) | Non-zero # of synthetic training images needed per 100 real images at max Test Accuracy | Max Test Accuracy (mixed) |
|---|---|---|---|
| Cat_Dog | | | |
| VGG16 | 96.4% | 2 | 96.7% |
| MobileNet | 91.9% | 5 | 91.8% |
| Weld | | | |
| VGG16 | 82.1% | 83 | 86.8% |
| MobileNet | 83.9% | 1 | 84.2% |

Fig. 4.2. As synthetic images are added to the training set, the test-training accuracy gap increases but it is unclear (other than the VGG16 Weld Defect case) that test accuracy increases.

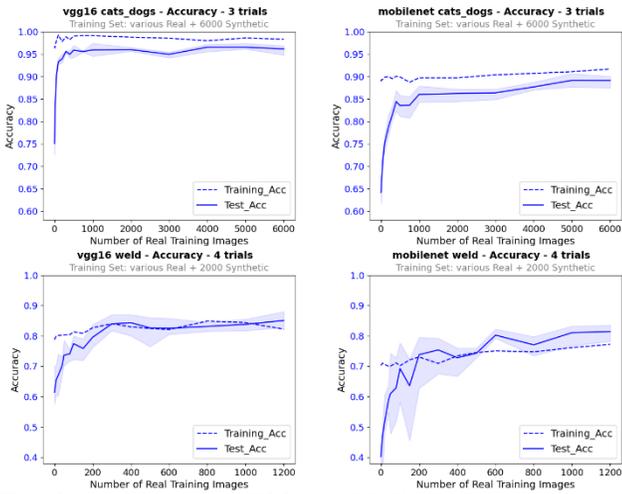

Fig. 4.3. Accuracy results of VGG16 and MobileNetV3-small models trained on increasingly added real images. As the model is trained on more real images, test accuracy increases. The blue cloud is the max/min test accuracy in collected samples.

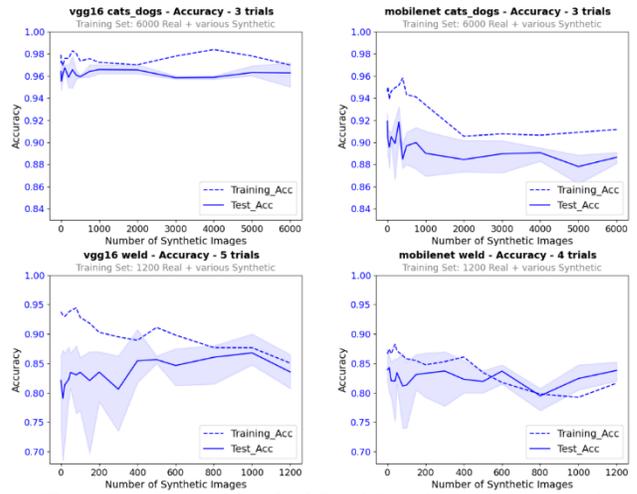

Fig 4.4. Accuracy results of VGG16 & MobileNet models trained on increasingly prevalent synthetic weld defect images. As the models are trained on more synthetic images, the test accuracy does not change consistently while the difference between testing & training accuracy increases.

### B. Quantifying Synthetic Image Quality (Cats_Dogs)

Regarding synthetic quality, all real (12,500/12,500) and all synthetic (20633/16726) cat & dog images are scored by a frozen 1000-class ImageNet weight pre-trained VGG16 and MobileNet-small models to better understand the differences between each image type's feature space. Each model and image type combination produced a histogram of confidence scores for images that the frozen model correctly classified in their top-3. All incorrectly classified images (when the real class is outside of the top-3) are not included in the histogram.

| Actual Class | 1st Top | 2nd Top | 3rd Top | Result |
|---|---|---|---|---|
| Cat | Persian Cat | Towel | Tubby Cat | Correct |
| Dog | Cloth | Shovel | Terrier | Correct |
| Cat | Rake | Grass | Towel | *Misclassified* |

Figure 4.5. Example top-3 results. Any image with a valid class in its top-3 is listed as correct and is included in the histograms.

This paper assumes that the shape of the confidence score distribution histogram can describe the real-synthetic gap for each class. This interpretation focuses on the differences in histogram distribution. If the histograms for real & synthetic images have similar mean, skew, and quartile shapes then the domain gap is believed to be small. As seen in Fig. 4.6 & 4.7, both model auditors show there is a significant difference in top-3 confidence and distribution shape between the real and synthetic cat & dog images.

The VGG16 model with frozen ImageNet weights correctly top-3 classified ~83.4% of the real cats & dog images, while only correctly classifying ~14.7% of the synthetic. As seen in the histograms below, the auditor's real image histogram appears like a normal distribution, where correct top-3 classifications are a mix of high & low confidence (even the ~300 images in the lowest confidence bin (<5%)). The confidence score distribution for the synthetic images are skewed heavily to the right and could not correctly classify most images (only 14.3% in top-3). This implies that there is a significant gap between real and synthetic cat & dog images. Further visual inspection of the images with the best & worst confidence scores can infer which features most contributed to the model's feature space of the cat or dog class.

Like VGG16, the MobileNetV3-small model, with frozen ImageNet weights, correctly top-3 classified ~88.3% of the real cats & dog images, while only correctly classifying ~15.3% of the synthetic. Compared to the VGG16 frozen model, MobileNetV3's distribution skewed more to the left (AKA: more correct classifications had a confidence score <50%) as seen in Fig. 4.7 below. The left skew is also observed in Fig. 4.6 for VGG16's synthetic histogram. In terms of accuracy, the MobileNetV3 auditor incorrectly classified most of the images. However, the smaller MobileNetV3-small model correctly classified more images than the larger VGG16model (while being less confident in its correct classifications than VGG16).

One benefit of auditing each of the dataset's images is that the auditor makes it possible to review why an image had a high/low confidence score and what it is misclassified as. The generated synthetic cats are mostly mistaken for items found in domestic settings (such as a chest, umbrella, curtain, etc), for smaller dog breeds (such as terriers and basenjis), and common items you might find around a cat (such as a chest, cloth, etc). These struggles imply that the models did not focus on the cat in the image or when the animal was found the model could not extract the unique features of cats from smaller dogs. Synthetic dogs are mostly mistaken for domestic items (such as brooms) or larger goats. Repeating these tests with a model architecture that support 'attention' may yield different results.

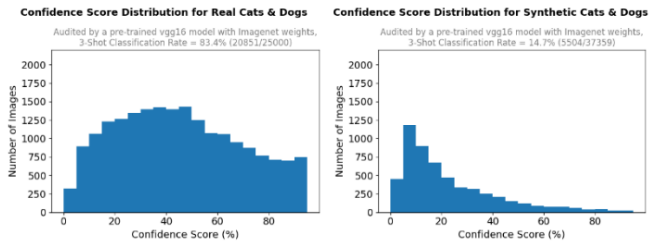

Fig. 4.6. Histogram of Confidence scores of top-3 classified (left) real and (right) synthetic Cats & Dogs images from a frozen VGG16 model pre-trained on ImageNet 1000-class weights. Images not top-3 classified are removed. This paper hypothesizes that these differently shaped curves imply that the model's extracted feature spaces for real & synthetic images are not similar.

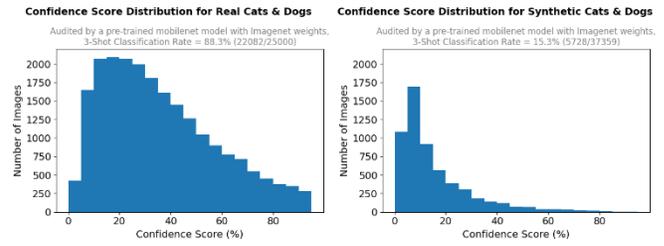

Fig. 4.7. Histogram of Confidence scores of top-3 classified (left) real and (right) synthetic Cats & Dogs images from a frozen MobileNetV3-small model pre-trained on ImageNet 1000-class weights. Images not top-3 classified are removed. This paper hypothesizes that these similar curves imply that the model's extracted feature spaces for real & synthetic images are closely aligned.

## C. Other experiments: Synthetic availability and rebalancing imbalanced datasets

Other uses of synthetic images were tested in addition to measuring the effect of mixed training sets and auditing the quality of images; 1) effect of synthetic availability and 2) possible benefits of rebalancing imbalanced datasets with synthetic images. First, the 'Increment Synthetic' test was repeated with differing amounts of available synthetic images to be sampled (9000 or ~4700 synthetic images). The difference between the two sets is the precision in which the synthetic images were generated (IE: 1 image every 5 degrees of rotation around a model versus every 30 degrees). The results showed that there was no significant change in model test accuracy when sampling from a larger dataset versus a smaller one. The second additional test used the gathered Weld Defect dataset to compare methods of rebalancing a training set with imbalanced classes. Cost-Learning, Undersampling, Oversampling, and Synthetic Padding methods were tested [13]. After training models using each method, results showed that Undersampling consistently provided the lowest test accuracy. However, the difference in test accuracy between the other three methods is minor for these cases.

## V. CONTRIBUTIONS TO THE FIELD

This work is relevant to anyone interested in improving their CV classification models. The audience includes production line managers looking to improve their quality control methodologies, engineers trying to improve model generalization for self-driving cars, or any other professionals wanting to create better classification models. This paper focuses on how training CV models on a mix of real & synthetic images can affect model accuracy and generalization. Secondly, the paper investigates other ways user-created synthetic images can be used. For example, pre-trained CV models may be used to judge the quality of synthetic images to further improve the benefits of training with synthetic data (aka. better define the real-synthetic domain gap).

## VI. FUTURE RESEARCH

A future experiment can continue this research by expanding the use of photo-realistic rendered images. First, re-attempt these experiments with synthetic images that include metadata (such as bounding boxes) with models that support state-of-the-art methods. This involves improving the image rendering pipeline to create images with the additional metadata. Second, continue experimenting with higher quality mesh models to produce higher-quality synthetic images. Finally, utilize the pre-trained auditor's judgment to further refine the eventual dataset used to train CV models. Possibly only training on the best synthetic images. In theory, if the 'worst quality' synthetic images (as by the pre-trained auditors) are removed, then the resulting subset of synthetic images will be of higher quality and be more likely to expand the feature space of the targeted class. Resulting in even greater accuracy and generalization benefits than using real images alone. We can conclude that it is acceptable to remove synthetic images, because our results suggest that only a minor number of synthetic images are needed to reach peak benefits.

## VII. CONCLUSIONS

These experiments explore how Deep CV models can benefit from training on synthetic images created in rendering engines like Unreal Engine. Even with a minimal number of created images based on low quality models (<10,000 polygon), the mix of training set images can improve the test-training accuracy gap and increase its peak test accuracy compared to training on just real or synthetic images alone. These effects can be seen for both large & small parameter Deep CV models. These results suggest that created synthetic images can expand a class's feature space by introducing examples not found in the training set or not yet feasible in the real world. In addition, pre-trained multi-class Deep CV models can act as an auditor to approximate the feature-space differences between the generated synthetic and real images, if the target class is included in the pre-trained weights, by comparing top-3 classification score distributions.

Flexibility. Generating synthetic data in visualization tools such as Epic Game's Unreal Engine (UE) can provide teams with a near infinite supply of images that replicate target classes in a controlled and repeatable environment. These virtual environments can be changed to follow new requirements without the need to re-collect massive datasets. They can even model situations that haven't been seen in your application, thus increasing the model's flexibility. The limit is your imagination and preparation.

Variation. The key to a successful CV classifier is having a dataset with a diverse set of features to learn from. These findings agree with previous research published by Dr. Thornström, Dr. Gastelum, and the anonymous ECCV submission stating that synthetic image variation is the key to training on synthesized images. Producing large quantities of synthetic images with small variations does not guarantee feature space expansion. Rather, images with significant variations contribute more to a model's classification performance. Based on these results, it is the author's opinion that synthetic image quality, rather than quantity, contributes to feature space expansion. This work, and the others cited, suggest that future developers may see more success if they focus on gathering varied images from different perspectives rather than capturing every statistical variation of an image. Proportional sampling of generated synthetic images, producing high quality & significantly different images, and using various UE pre-built environments for synthetic image capture may increase synthetic image effectiveness.

### A. Analysis of the real synthetic domain gap

By analyzing which synthetic cat and dog images have the lowest pre-trained model confidence scores, a developer can understand which synthetic models, backgrounds, and augmentations the model recognizes the best. If a limited number of synthetic mesh models are used, it is possible to correlate a synthetic mesh model to a specific ImageNet class, thus allowing a developer to filter which synthetic images align best to their given application, potentially further increasing the advantages of training on that subset of synthetic images. For example, 18 of the 20 most confident top-1 correct predictions of synthetic cat images for VGG16 were of an Egyptian Cat or Lynx. For synthetic dogs, 12 out of the top 14 most confident VGG16 classifications were of Great Danes.

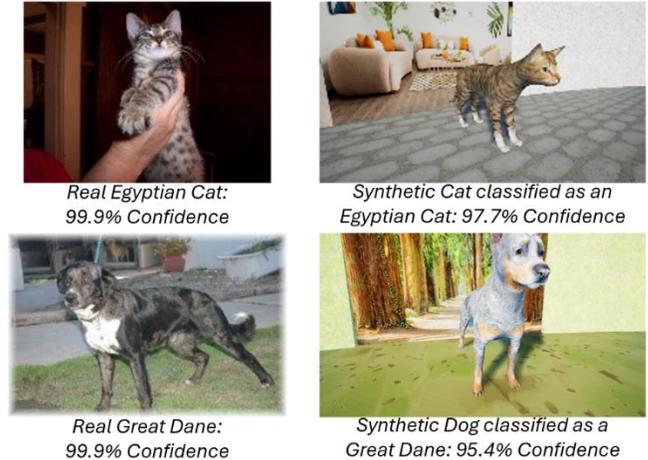

Fig. 7.1. Comparison of correct Top-1 classifications for VGG16 for Persian Cat and Great Dane images. Top-1 Confidence scores for each are above 99.7%. Infers that there is a relationship with the real breeds and the Unreal Engine mesh model they are simulating.

When looking at the images with the best & worst confidence scores by the Cat_Dog auditor (Fig. 7.2), no one variation consistently indicated a good/bad image, but they did share similar characteristics. High confidence correctly classified images had a zoom fit on the target, balanced lighting, and the target centered in the frame. Low confidence correctly classified images contained overly saturated or dim/poor lighting with highly complex background patterns where the target was hard to pick out.

One surprise is that the smaller model (MobileNetV3-small) had a higher top-3 classification rate for both real and synthetic images compared to VGG16. MobileNetV3 correctly classified +4.9% more real cat & dog images and +0.6% more synthetic images compared to VGG16. It is possible that the lower parameter count allowed the pre-trained ImageNet weights to not overfit to the features of real images as much as the larger VGG16 model. Optimally, both models would show similar distributions and classification rates to exhibit a consistent opinion of the real-synthetic domain gap. One model may not concentrate on the same features as the other. Regardless, the different auditor performance and distributions are an example of how different model architectures form different feature spaces for the same object. The models still can be used to gain insights into what makes the best or worst images for a class.

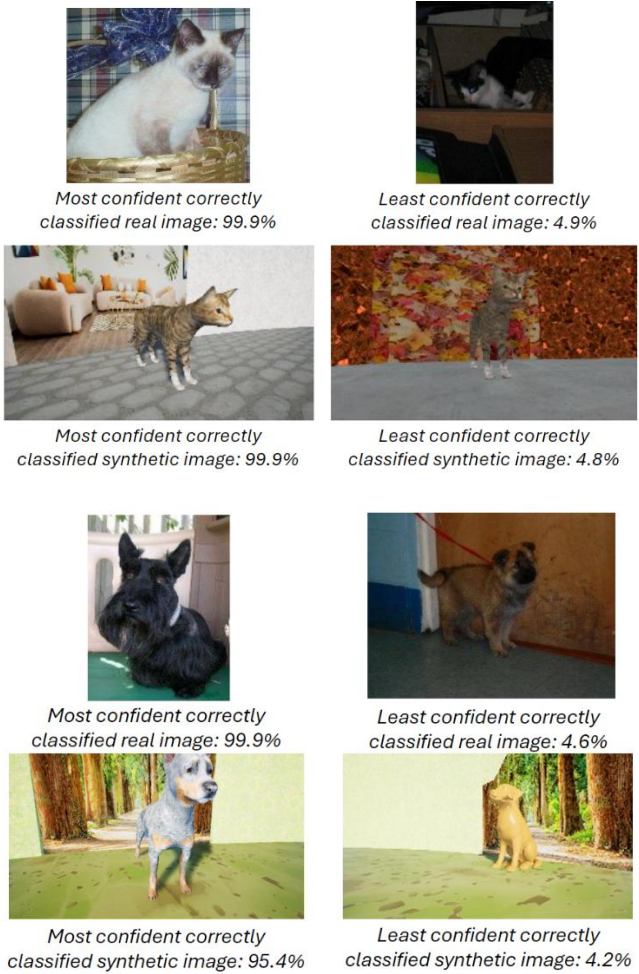

Fig. 7.2. Examples of high & low confidence cat dog images.

### B. Advice on making synthetic images

This paper makes the following recommendations to improve generated synthetic data. First, utilize randomness in every facet of 'setting up' a screenshot (like the Welding collection), instead of collecting images on a set track with hardcoded values for each image (like the Cat_Dog collection). Note: these results suggest that higher-polygon UE models contribute more to the class feature space than taking more images from slightly different angles. Second, developers should proportionally sample the rendered images to ensure all meta-data values are included (model, lighting, etc). Proportionally sampling can help prevent oversampling of one characteristic over another to maintain image variation. Lastly, try incorporating more expansive UE environments during image collection. Only the UE test environment is used in this experiment, there are fully fleshed out landscapes, cities, and environments ready to insert the desired 3D models into.

### C. Commentary on using GANs for image generation

Regarding the use of GANs in synthetic image creation, as of the date of publication, open-source tools can produce highly detailed images with user provided text prompts that can be locally saved. In-house developed GANs can be trained to provide slight variations, such as the ECCV paper's use to improve the image 'realism' of created synthetic images [3]. This is an alternative to this report's process of creating a UE environment then setting up the pipeline to generate synthetic images. It is recommended that developers understand the workflow and business strengths and risks of using GANs or UE environments for their image pipelines.

GANs produce images by selecting from learned distributions of features, thus creating pseudo-random images that are difficult to recreate (once trained, fast, but low reproducibility and control). If a third-party GAN is used, consider how to retain images and create more if the API were to suddenly change (see API changes for Reddit and Twitter in 2023). Image pipelines based on UE environments allows users to reproduce images using the exact variations that they want and offers a nimble way of generating new images as customer requirements change (fast & high reproducibility but requires prior knowledge of rendering engines).

## VIII. FOR MORE INFORMATION

Please send an email to *smutnyj@vt*.edu for any questions or source code requests related to this report. This includes anything related to implementing Unreal Engine environments, and/or UE python callbacks.